\documentclass{article}
\usepackage{spconf,amsmath,graphicx}
\usepackage{amsthm,amsmath,amssymb}
\usepackage{mathrsfs}
\usepackage{booktabs}
\usepackage{threeparttable}
\usepackage{multirow,multicol}
\usepackage{makecell}
\usepackage{bm}

\title{Improving CTC-based speech recognition via knowledge transferring from pre-trained language models}
\name{Keqi Deng$^{2,3,*}$, Songjun Cao$^{1,*}$, Yike Zhang$^1$, Long Ma$^1$, Gaofeng Cheng$^{2}$, Ji Xu$^{2,}$\textsuperscript{†}, Pengyuan Zhang$^{2,3}$ 
\thanks{ $^*$ Equal contribution. \textsuperscript{†} Corresponding author.}
\thanks{This work is partially supported by the National Key Research and Development Program of China (No. 2020AAA0108002).}}
\address{
  $^1$Tencent Cloud Xiaowei, Beijing, China\\
  $^2$Key Laboratory of Speech Acoustics and Content Understanding, Institute of Acoustics, CAS, China\\
  $^3$University of Chinese Academy of Sciences, China}
\begin{document}
\ninept
\maketitle
\begin{abstract}
Recently, end-to-end automatic speech recognition models based on connectionist temporal classification (CTC) have achieved impressive results, especially when fine-tuned from wav2vec2.0 models.
Due to the conditional independence assumption, CTC-based models are always weaker than attention-based encoder-decoder models and require the assistance of external language models (LMs).
To solve this issue, we propose two knowledge transferring methods that leverage pre-trained LMs, such as BERT and GPT2, to improve CTC-based models.
The first method is based on representation learning, in which the CTC-based models use the representation produced by BERT as an auxiliary learning target. 
The second method is based on joint classification learning, which combines GPT2 for text modeling with a hybrid CTC/attention architecture.
Experiment on AISHELL-1 corpus yields a character error rate (CER) of $4.2\%$ on the test set.
When compared to the vanilla CTC-based models fine-tuned from the wav2vec2.0 models, our knowledge transferring method reduces CER by $16.1\%$ relatively without external LMs.
\end{abstract}
\section{Introduction}
\label{sec:intro}
End-to-end (E2E) automatic speech recognition (ASR) models directly transcribe input speech into corresponding transcripts and outperform conventional ASR models like hidden Markov models (HMMs) on most public corpora \cite{8682586,9414594}.
Among E2E models, the attention-based encoder-decoder (AED) model is
the dominant structure \cite{8682586,9398531}. 
However, the AED models' outstanding performance comes at a cost of large model size and computational complexity \cite{9414594}.
The AED models, in particular, decode in an auto-regressive (AR) fashion, which slows down the decoding speed when beam size is large \cite{DBLP:journals/corr/abs-2104-04805}.
In contrast, connectionist temporal classification (CTC)-based models have a conditional independence assumption, allowing for fast and paralleled decoding \cite{9414594}.

However, although the self-supervised pre-training methods have achieved impressive success and greatly improve the recognition performance of CTC-based ASR models \cite{Schneider2019,DBLP:conf/iclr/BaevskiSA20,NEURIPS2020_92d1e1eb}, it is usually weaker than the AED-based ASR system due to its conditional independence assumption \cite{9414594}. Using an external AR LM can help alleviate this problem \cite{8462576,deng21b_interspeech}, but it is more meaningful and flexible to improve the performance of the CTC-based model itself, because using the AR LM during beam search makes the system an AR one and loses the advantage of fast decoding speed as a non-autoregressive (NAR) model.
The recently proposed Mask CTC \cite{Higuchi2020} and Imputer \cite{pmlr-v119-chan20b} employ the CTC as part of the NAR ASR system and try to refine the prediction of the CTC part. But these methods demand extra computational costs and additional parameters.

The CTC-based models are limited in utilizing text-modal contextual information due to its conditional independence assumption \cite{9414594}. We aim to help the CTC-based models learn the text-modal knowledge without introducing extra parameters to keep the overall structure simple during inference. Therefore, we try to transfer the contextual knowledge of those well-known pre-trained LMs like BERT \cite{DevlinCLT19} or GPT2 \cite{radford2019language} to the CTC-based ASR models. In this paper, we employ wav2vec2.0 \cite{NEURIPS2020_92d1e1eb} as the vanilla CTC-based ASR system given its recent state-of-the-art results and propose two knowledge transferring methods to further improve its performance. The first method is based on representation learning, in which the CTC-based ASR system employs the representation extracted by the BERT \cite{DevlinCLT19} as an auxiliary learning target.
During this process, we distill the contextual knowledge into the CTC-based ASR system. The second method is based on joint classification learning that borrows the ideas of CTC/attention multi-task learning method \cite{kim2017multi}. 
We combine the GPT2 for text modeling with a hybrid CTC/attention architecture, 
and the cross-modal knowledge transferring is achieved during the joint training.
Experimental results show that the first method performs better since it transfers
bidirectional contextual knowledge into the ASR system, and it achieves a $4.2\%$ CER on the test set of AISHELL-1 corpus \cite{8384449}.
Without an external LM, our proposed method yields $16.1\%$ relative CER reduction
compared with our strong wav2vec2.0 CTC baseline and can still outperform the AED-based ASR systems on the AISHELL-1 benchmark.

The rest of this paper is organized as follows. In Section~\ref{related}, we introduce the related works. In Section~\ref{proposed}, we describe the proposed knowledge transferring methods. The experiments and conclusions are presented in Sections~\ref{sec:experiments} and \ref{sec:con}, respectively.\par

\section{Related works}
\label{related}
Self-supervised pre-training has gained success in several
fields like natural language processing (NLP) \cite{DevlinCLT19,radford2019language}, computer vision (CV) \cite{pmlr-v119-chen20j} and ASR \cite{NEURIPS2020_92d1e1eb}.
Recently well-known BERT contains multi-layer Transformer encoders and is pre-trained via masked learning \cite{DevlinCLT19}. Unlike BERT, GPT2 \cite{radford2019language} consists of a stack of unidirectional Transformer decoders and is well known as a language generator \cite{DBLP:journals/corr/abs-2004-02251}.

Inspired by BERT, wav2vec \cite{Schneider2019} learns the raw speech representations via a self-supervised context-prediction task.
And vq-wec2vec \cite{DBLP:conf/iclr/BaevskiSA20} is designed to learn discrete speech representations.
Finally, wav2vec2.0 combines the contextual representations and discrete speech units learning into one system and can outperform previous works with much less labeled data \cite{NEURIPS2020_92d1e1eb}.

Many works have investigated to transfer the knowledge of BERT to the AED-based ASR system \cite{9437636, 9664007}. Futami et al. \cite{futami2020distilling} and Liu et al. \cite{DBLP:conf/icassp/LiuSCLL20} distill the knowledge into a sequence-to-sequence (S2S) ASR system from the BERT via soft labels and multi-objective function, respectively. Winata et al. transfer part of the BERT's parameters to the decoder of the S2S ASR models \cite{DBLP:journals/corr/abs-2012-01687}. 
Since the S2S model itself has some abilities to utilize text-modal information, these methods have achieved limited improvements. Actually, transferring the contextual knowledge of pre-trained LMs to the CTC-based ASR systems is more effective since they do not contain an auto-regressive decoder to utilize text modality. 

\section{Proposed method}
\label{proposed}
Under the framework of wav2vec2.0 \cite{NEURIPS2020_92d1e1eb}, we propose two knowledge transferring methods to further improve the CTC-based models. The first is based on representation learning using the BERT \cite{DevlinCLT19} and we refer to it as KT-RL in the rest of this paper.
The second method is based on joint classification learning utilizing GPT2 \cite{radford2019language} and we refer to it as KT-CL.
\subsection{Knowledge transferring based on representation learning}
\begin{figure}[h]
    \centering
    \includegraphics[width=63mm]{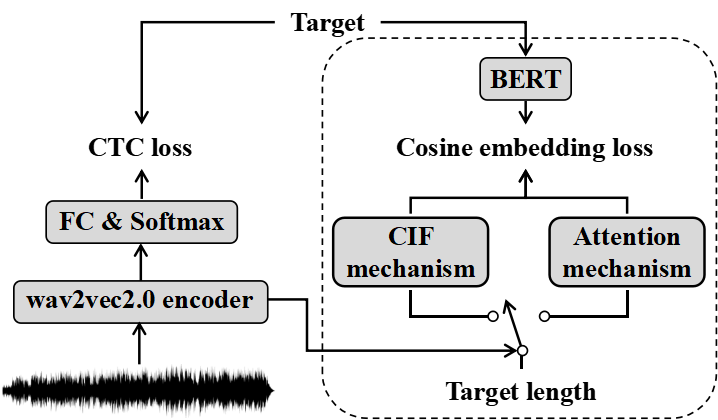}
    \caption{Illustration of proposed knowledge transferring methods based on representation learning. }
    \label{bert}
\end{figure}

Our method is shown in Fig.~\ref{bert}, where the switch is independent on the target length and
FC denotes a fully connected layer.
The wav2vec2.0 encoder contains a CNN-based feature encoder and a Transformer-based network. The parameters of BERT are fixed.

Suppose the representation vectors extracted by wav2vec2.0 encoder as $\textbf{H}=({\bm{h}}_1, \dots, {\bm{h}}_M)$ and the corresponding target labels as $\textbf{T}=({y}_1, \dots, {y}_N)$. The CTC loss $\mathcal{L}_{\rm{ctc}}$ is calculated between the $\textbf{T}$ and the $\textbf{H}$ after the FC layer. To transfer the knowledge of BERT into the ASR system, we should first extract a linguistic representation whose time dimension is consistent with the target $\textbf{T}$. Inspired by the work \cite{9398531}, we use the serial CIF mechanism \cite{9054250} to achieve the monotonic alignment between the speech and text modalities. In addition, we design another structure based on attention mechanism \cite{Vaswani2017} to achieve better parallel efficiency.
\subsubsection{KT-RL based on CIF (KT-RL-CIF)}
Inspired by the work \cite{9398531}, 
we apply a sigmoid function to the max element of $\bm{h}_m$ after a FC layer to obtain the weight $w_m$. And the $w_m$ are accumulated along the time dimension and a linguistic representation $\bm{l}_n$ is output when accumulated resized weight $\hat{w}_m$ surpasses 1. The specific process is shown as follows:
\begin{equation}
    {w_m} = {\rm sigmoid}({\rm max}({\rm FC}(\bm{h}_{m}))),\label{a}
\end{equation}
\begin{equation}
    {\hat{w}_m} = \frac{w_m}{\sum_{m=1}^{M}w_m}N,\label{what}
\end{equation}
We input the $\hat{w}_m$ and $\bm{h}_{m}$ into CIF \cite{9054250}, which accumulates the $\hat{w}_m$ and integrates the $\bm{h}_{m}$ (in weighted sum form) until the accumulated $\hat{w}_m$ surpasses 1. At this time step, the CIF divides $\hat{w}_m$ into two part: one part is to fill the accumulated $\hat{w}_m$ to 1; the other part corresponds to the next linguistic representation. Then, the CIF fires the extracted linguistic representation $\bm{l}_n$.
Eq.~\ref{what} is to ensure that the number of output $\bm{l}_n$ is consistent with the time length of the target $\textbf{T}$. The process based on CIF extracts a linguistic representation $\bm{l}_n$ based on monotonic alignment.

We input the target transcription into the BERT and output the extracted contextual embedding $\textbf{E}=({\bm{e}}_1, \dots, {\bm{e}}_N)$. We employ the $\bm{e}_n$ as the learning target of the $\bm{l}_n$, through which we distill the knowledge of BERT into the ASR system. We calculate the cosine embedding loss between the $\bm{e}_n$ and $\bm{l}_n$:
\begin{equation}
     \mathcal{L}_{cos} = k\cdot\sum_{n=0}^N(1-{\rm cos}(\bm{l}_{n},\bm{e}_{n})), \label{cos}
\end{equation}
where $k$ is a hyper-parameter given that the value of cosine embedding loss is relatively small compared with other loss functions.

The final training objective $\mathcal{L}_{mtl}$ is calculated as follows:
\begin{equation}
    \mathcal{L}_{mtl}=\lambda \mathcal{L}_{\rm{ctc}} + (1\!-\!\lambda)\mathcal{L}_{\rm{cos}},
    \label{mot}
\end{equation}
where the hyper-parameters $\lambda \in (0,1)$. During inference, we only use the CTC-based structure for decoding.
\subsubsection{KT-RL based on attention (KT-RL-ATT)}
To achieve better parallel efficiency, we propose to use an attention-based structure to extract the linguistic representation.
With a prior known target length $N$, we get a positional embedding matrix $\textbf{P}=({\bm{p}}_1, \dots, {\bm{p}}_N)$ after positional encoding and broadcasting the dimension to be consistent with the wav2vec2.0 encoder output $\textbf{H}$.

After that, the linguistic representation $\textbf{L}=({\bm{l}}_1, \dots, {\bm{l}}_N)$ is got via a parallel multi-head cross attention:
\begin{equation}
    \bm{head}_{i} = {\rm softmax}\left( {\frac{(\textbf{P}\textbf{W}_{i}^{Q})(\textbf{H}\textbf{W}_{i}^{K})^{T}}{\sqrt{d_{m}}}} \right)(\textbf{H}\textbf{W}_{i}^{V}), \label{head}
\end{equation}
\begin{equation}
    \textbf{L} = {\rm concat}\left( {\bm{head}_{1},\ldots,\bm{head}_{h}} \right)\textbf{W}^{O}, \label{com}
\end{equation}
where $\textbf{W}_{i}^{Q}\in\mathbf{R}^{d_{{model}} \times d_m}$, $\textbf{W}_{i}^{K}\in\mathbf{R}^{d_{{model}} \times d_m}$, and $\textbf{W}_{i}^{V}\in\mathbf{R}^{d_{{model}} \times d_m}$ are projection matrices for queries (Q), keys (K), and values (V) respectively. $d_{{model}}$ is the model dimension. And we calculate the $\mathcal{L}_{cos}$ and $\mathcal{L}_{mtl}$ in the same way as Eq.~\ref{cos} and Eq.~\ref{mot}.


\subsection{Knowledge transferring based on classification learning}
\begin{figure}[h]
    \centering
    \includegraphics[width=55mm]{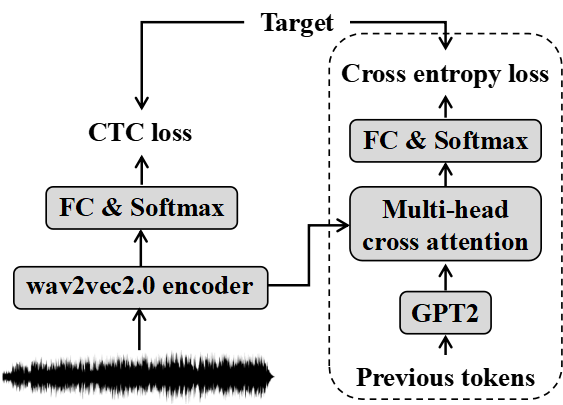}
    \caption{Illustration of proposed knowledge transferring methods based on joint classification learning. }
    \label{gpt2}
\end{figure}

Inspired by the CTC/attention learning methods \cite{kim2017multi} and Preformer \cite{deng2021improving}, 
we first use a GPT2 to extract a text-modal representation of previous tokens (ground truth) and combine it with the speech-modal encoder output via multi-head cross attention. The process of our method is shown in Fig.~\ref{gpt2}, where the parameters of GPT2 are fixed.

Suppose the embedding matrix that is extracted by the GPT2 from the previous tokens as $\textbf{G}=({\bm{g}}_1, \dots, {\bm{g}}_N)$. The process of multi-head cross attention is similar to the Eq.~\ref{head} and Eq.~\ref{com}, except that the matrix $\textbf{G}$ is employed as query matrix instead of $\textbf{P}$ and a subsequent mask \cite{Vaswani2017} is used to prevent seeing future tokens. During this process, the speech-modal encoder output $\textbf{H}$ have to learn to bridge the modal gap with the text-modal GPT2 output $\textbf{G}$ as shown in Eq.~\ref{head}, in this way, the text-modal knowledge is transferred into the CTC-based ASR system.

Suppose the cross attention output as $\textbf{O}=({\bm{o}}_1, \dots, {\bm{o}}_N)$, we calculate the cross entropy loss $\mathcal{L}_{ce}$ between the $\textbf{O}$ after linear classifier and the target.
And we train the overall structure via joint classification with the
final training objective $\mathcal{L}_{mtl}$ calculated as follows:
\begin{equation}
    \mathcal{L}_{mtl}=\beta \mathcal{L}_{\rm{ctc}} + (1\!-\!\beta)\mathcal{L}_{\rm{ce}},
    \label{mot2}
\end{equation}

\section{Experiments}
\label{sec:experiments}

\subsection{Corpus}
\label{ssec:corpus}
We evaluate the proposed knowledge transferring methods on the Mandarin AISHELL-1 corpus \cite{8384449},
which consists of about 120000 utterances for the training set and about 7000 utterances for the testing set. And its development set contains around 14000 utterances.
We also pre-train a Mandarin wav2vec2.0 base model using the unlabeled speech training set of AISHELL-2 corpus \cite{DBLP:journals/corr/abs-1808-10583}.

\subsection{Model descriptions}
\label{ssec:model description}
We use the ESPnet2 toolkit \cite{watanabe2018espnet} to build the models. For the acoustic input feature, we use the raw speech following wav2vec 2.0 \cite{NEURIPS2020_92d1e1eb}. For the text output, we use 4230 Chinese characters and 3 non-verbal symbols: blank, unknown-character, sos/eos
as the modeling units.

We employ the vanilla wav2vec2.0 CTC-based model as the baseline, which contains a wav2vec2.0 encoder and a randomly initialized FC.
The wav2vec2.0 encoder consists of seven CNN layers (i.e., 512 channels with kernel size 10, 3, 3, 3, 3, 2, and 2 and strides 5, 2, 2, 2, 2, 2, and 2) and 12 Transformer layers (i.e., 768 model dimensions, 3072 inner dimensions (FFN) and 12 attention heads). 
As for our proposed methods, except for the CTC-based structure that is the same as the baseline model,
we set the $k$ in Eq.~\ref{cos} to 20, and the $\lambda$ and $\beta$ in Eq.~\ref{mot} and Eq.~\ref{mot2} to 0.3. The multi-head cross attention mechanism in the proposed methods is only one layer with 768 model dimensions and 4 heads. And the sizes of the FC in 
Eq.~\ref{a}, Fig.~\ref{bert}, and Fig.~\ref{gpt2}
are all 4233.
We directly use the pre-trained BERT (i.e., bert-base-chinese) and GPT2 (i.e., uer/gpt2-chinese-cluecorpussmall) provided by Huggingface Transformer Library \cite{wolf-etal-2020-transformers}. The parameters of both the GPT2 and the BERT are fixed during training.
The averaged embedding over all layers of BERT is employed as the target of KT-RL.
We also use the wav2vec2.0 base model provided by Fairseq \cite{ott2019fairseq} as the English wav2vec2.0 model.
 
We train the models by Adam optimizer with a warmup learning rate schedule (25000 warm steps) for 20 epochs. 
To avoid overfitting,
we use an early stop strategy and the patience is set to 3. And we average the parameters of the model at the last 10 epochs.
We also train an external Transformer-based LM using the transcription of the training set of AISHELL-1 following ESPnet2 recipe \cite{watanabe2018espnet}. During inference, we only use the CTC-based model for decoding, and the beam search size is 10. When the external LM is used, its weight is set to 0.3.

\subsection{Experimental results}
We compare our proposed knowledge transferring method, our vanilla wav2vec2.0 CTC baseline, and other ASR benchmark systems. The results are shown in Table~\ref{tab:PC}, where w2v2.0 denotes the mandarin wav2vec2.0 base model. 
The results show that our proposed method greatly improves the CTC-based models, especially when not using the external LM (i.e., $14.3\%$ and $16.1\%$ relative CER reduction on the test set when using LM and not using LM respectively). These results prove that transferring the contextual knowledge of powerful pre-trained LM to the CTC-based ASR system is really effective since the CTC-based models lack the ability to utilize the text-modal information. 

In addition, compared with those benchmark systems, we can see that the CTC-based models after using our proposed method achieve
very competitive results,
even though we
greatly reduce the number of training iterations. For example,
the Conformer with SpecAugment \cite{Park2019} in ESPnet2 employs 50 training epochs, but we just train our model for 20 epochs, 
which greatly saves training costs. 

Furthermore, with our proposed knowledge transferring method, the CTC-based non-autoregressive ASR system without an external autoregressive (AR) LM can still surpass our strong wav2vec2.0 CTC baseline with an AR LM, which can be seen as an AR system.
\begin{table}[h]
  \caption{The character error rates (CERs) (\%) of several benchmarks, our vanilla wav2vec2.0 CTC-based baseline, and our proposed  knowledge transferring methods on the AISHELL-1 corpus.}
  \setlength\tabcolsep{3.5pt}
  \label{tab:PC}
  \centering
  \begin{tabular}{l c c c c}
    \toprule
    \multirow{2}{*}{ASR Model} &
    \multicolumn{2}{c}{With LM} & \multicolumn{2}{c}{No LM}\\
     & dev & test & dev & test\\
    \midrule
    ESPnet2 (Transformer+SpecAug \cite{Park2019}) \cite{watanabe2018espnet} & 5.9 &6.4&--&--\\
    ESPnet2 (Conformer+SpecAug \cite{Park2019}) \cite{watanabe2018espnet} & 4.4 &4.7&4.5&4.9\\
    Improved CASS-NAT \cite{fan21b_interspeech} & --&--&4.9&5.4\\
    NAR-BERT-ASR \cite{DBLP:journals/corr/abs-2104-04805} & --&--&4.9&5.5\\
    LASO \cite{9437636} & --&--&5.2&5.8\\
    \midrule
    Vanilla w2v2.0 CTC &4.5&4.9&5.1&5.6\\
    KT-RL-CIF based on w2v2.0 &\textbf{4.1}&\textbf{4.2}&\textbf{4.3}&\textbf{4.7}\\
    \bottomrule
  \end{tabular}
\end{table}

\subsection{Ablation studies on KT-RL}
In the main experiments, we only show the performance of the proposed KT-RL-CIF. In this part, we conduct ablation studies to compare the KT-RL-CIF and the KT-RL-ATT. In addition, we employ the cosine embedding loss as in Eq.~\ref{cos} for the representation learning, in this part, we also compare the effectiveness of cosine embedding loss with the mean square error (MSE) loss. 

The results are shown in Table~\ref{repre}, where 
Aux Loss means an auxiliary loss in Eq.~\ref{cos}, and Cosine denotes cosine embedding loss while MSE denotes MSE loss. 
We can see that 
our proposed knowledge transferring methods based on representation learning can greatly improve the CTC-based models. Among them, KT-RL-CIF is better as the CIF constrains a monotonic alignment between acoustic and linguistic representation. Using the CIF mechanism can more accurately transfer the context knowledge to the corresponding acoustic representation. While the attention mechanism is too flexible and extracts the linguistic representation through a weighted sum, so knowledge corresponding to a word may be transferred to other frames that do not belong to this word in a certain proportion. In addition, the results show that using the cosine embedding loss outperforms the MSE loss.
The cosine embedding loss encourages the linguistic representation to approach the direction of the target embedding instead of trying to reduce the distance between each element of the two vectors. 
And the results imply that angle may be more important than the norm of a vector during representation learning.

\begin{table}[h]
  \caption{The CERs (\%) of our ASR system with different structures for knowledge transferring based on representation learning on the AISHELL-1 corpus.}
  \setlength\tabcolsep{4pt}
  \label{repre}
  \centering
  \begin{tabular}{l c c c c c}
    \toprule
    \multirow{2}{*}{ASR Model} & Aux&
    \multicolumn{2}{c}{With LM} & \multicolumn{2}{c}{No LM}\\
   & Loss & dev & test & dev & test\\
    \midrule
   Vanilla w2v2.0 CTC&-- &4.5&4.9&5.1&5.6\\
    \midrule
    KT-RL-CIF based on w2v2.0&Cosine &\textbf{4.1}&\textbf{4.2}&\textbf{4.3}&\textbf{4.7}\\
     KT-RL-CIF based on w2v2.0&MSE &4.4&4.7&4.8&5.1\\
    KT-RL-ATT based on w2v2.0&Cosine &4.2&4.5&4.6&4.8\\
    \bottomrule
  \end{tabular}
\end{table}

However, although the KT-RL-CIF outperforms the KT-RL-ATT in recognition accuracy, the KT-RL-CIF's training time will be longer due to its serial structure, while the KT-RL-ATT has a higher parallel efficiency thus shortening the training duration. 

\begin{table}[h]
  \caption{The training time spent in each iteration of different structures for knowledge transferring based on representation learning 
  on the AISHELL-1 corpus.}
  \setlength\tabcolsep{5pt}
  \label{time}
  \centering
  \begin{tabular}{l c c c}
    \toprule
    ASRmodel&{Forward Time} & Backward Time&Total Time\\
   \midrule
   Baseline& 0.171(s/iter)& 0.185(s/iter)& 0.356(s/iter)\\
   KT-RL-ATT&0.196 (s/iter)&0.263 (s/iter)&0.459 (s/iter)\\
   KT-RL-CIF&0.305 (s/iter)&0.414 (s/iter)&0.719 (s/iter)\\
    \bottomrule
  \end{tabular}
\end{table}
Table~\ref{time} lists the forward, backward, and total time of the KT-RL-ATT and KT-RL-CIF for each iteration respectively, which are trained on 8 M40 GPUs with 128 batch size.
We can see that the KT-RL-ATT does greatly reduce the time spent on training. To conclude, due to the monotonic alignment achieved by the CIF mechanism,
the KT-RL-CIF can better transfer the BERT knowledge to the CTC-based ASR system.
But the KT-RL-ATT can save the time spent on training while also greatly improve the ASR system.
\subsection{Ablation studies on KT-CL}
In this part, we conduct ablation studies to show the performance of the knowledge transferring based on joint classification. And the results are shown in Table~\ref{class}, where joint decoding means joint CTC/attention decoding \cite{CTC-Attention-ACL-2017}. Since the proposed KT-CL method is inspired by the CTC/attention multi-task learning, we build a vanilla CTC/attention ASR model as another baseline,
in which we employ CTC/attention joint training and only use the CTC branch during decoding. 
The results show that the CTC-based ASR system can be improved even with the vanilla CTC/attention multi-task training.

Furthermore, after using the proposed KT-CL, further improvements are achieved. 
The GPT2 can extract a high-quality text representation that fully contains contextual knowledge,
and the acoustic encoder output has to match this text-modal representation during joint training thus learning the contextual knowledge. And another advantage of the KT-CL is that we can choose another decoding style (i.e., joint CTC/attention decoding) if necessary, which can achieve higher accuracy although the inference speed is slower. 

At last, we can see that the KT-RL outperforms the KT-CL, this is because KT-RL distills bidirectional contextual knowledge into the ASR system while the KT-CL can only provide unidirectional information as it needs to use a subsequent mask to prevent seeing the future information during joint training.

\begin{table}[h]
  \caption{The CER (\%) of our ASR system with different structures for knowledge transferring based on classification learning on the AISHELL-1 corpus.}
  \setlength\tabcolsep{4pt}
  \label{class}
  \centering
  \begin{tabular}{l c c c c}
    \toprule
    \multirow{2}{*}{ASR Model} & 
    \multicolumn{2}{c}{With LM} & \multicolumn{2}{c}{No LM}\\
    & dev & test & dev & test\\
    \midrule
   Vanilla w2v2.0 CTC &4.5&4.9&5.1&5.6\\
    \midrule
    KT-CL based on w2v2.0 (joint decoding)
    &\textbf{4.0}&\textbf{4.3}&\textbf{4.0}&\textbf{4.2}\\
    \midrule
    KT-CL based on w2v2.0 &4.4&4.6&5.0&5.2\\
    CTC/attention based on w2v2.0 &4.6&4.8&5.0&5.4\\
    \bottomrule
  \end{tabular}
\end{table}
\subsection{Ablation studies on pre-trained LM}
In the previous experiments, we have demonstrated the effectiveness of our proposed KT-RL based on BERT and KT-CL based on GPT2. In this part, we further conducted ablation studies to test the effect of using different pre-trained LMs. The results are shown in Table~\ref{encoder}, where unidirectional BERT is achieved by subsequent masks.
The experimental results show that: 1. KT-RL-CIF with BERT outperforms the one with GPT2, which proves that using bidirectional information is beneficial for the KT-RL. 2. For the KT-CL, the effect of setting BERT to unidirectional is just slightly worse than GPT2. 3. Whether using GPT2 or BERT, the KT-RL outperforms the KT-CL.
\begin{table}[h]
  \caption{The CER (\%) of our ASR system using different pre-trained LM on the AISHELL-1 corpus, which is decoded with external LM.}
  \setlength\tabcolsep{4pt}
  \label{encoder}
  \centering
  \begin{tabular}{l c c}
    \toprule
    {ASR Model} & dev & test \\
    \midrule
   Vanilla w2v2.0 CTC &4.5&4.9\\
    \midrule
    KT-RL-CIF based on w2v2.0 (using BERT) &\textbf{4.1}&\textbf{4.2}\\
    KT-RL-CIF based on w2v2.0 (using GPT2)&4.2&4.5\\
    KT-CL based on w2v2.0 (using GPT2)&4.4&4.6\\
    KT-CL based on w2v2.0 (using unidirectional BERT)&4.4&4.7\\
    \bottomrule
  \end{tabular}
\end{table}
\section{Conclusion}
\label{sec:con}
In this paper, we aim to improve the CTC-based models by transferring contextual knowledge of pre-trained LMs into the ASR systems and we propose two kinds of knowledge transferring methods. The first is based on representation learning, which employs the representation extracted by BERT as the auxiliary learning target of the CTC-based models and can distill bidirectional contextual knowledge into the ASR systems. The second method is based on joint classification learning, which combines GPT2 for text modeling with a hybrid CTC/attention structure and can utilize unidirectional contextual knowledge.
We conduct experiments on AISHELL-1 corpus and achieve a $4.2\%$ CER on the test set. Without the help of external LM, our proposed method yields $16.1\%$ relative CER reduction compared with our wav2vec2.0 CTC baseline and can still surpass the AED-based ASR systems on the AISHELL-1 benchmark.
%

\pagebreak

\bibliographystyle{IEEEbib}
\bibliography{strings,refs}

\end{document}